%% file: PaperForReview.tex
\documentclass[10pt,twocolumn,letterpaper]{article}

\usepackage{cvpr}              

\usepackage{graphicx}
\usepackage{amsmath}
\usepackage{amssymb}
\usepackage{booktabs}

\usepackage{multirow}
\usepackage{amsfonts}
\usepackage{siunitx}
\usepackage{pifont}
\usepackage[normalem]{ulem} 

\usepackage{fontawesome5}
\usepackage[pagebackref,breaklinks,colorlinks]{hyperref}
\usepackage{soul}

\usepackage{enumitem}

\usepackage[capitalize]{cleveref}
\crefname{section}{Sec.}{Secs.}
\Crefname{section}{Section}{Sections}
\Crefname{table}{Table}{Tables}
\crefname{table}{Tab.}{Tabs.}

\newcommand{\scotch}{{\fontfamily{cmtt}\selectfont SCOTCH}}
\newcommand{\soda}{{\fontfamily{cmtt}\selectfont SODA}}

\def\cvprPaperID{9653} 
\def\confName{CVPR}
\def\confYear{2023}

\newcommand{\xmark}{\ding{55}}%

\begin{document}

\title{SCOTCH and SODA: A Transformer Video Shadow Detection Framework}

\author{Lihao Liu$^{1}$,\: Jean Prost$^{2}$,\: Lei Zhu$^{3,4}$,\: Nicolas Papadakis$^{2}$,\: Pietro Liò$^{1}$,\: \\ Carola-Bibiane Schönlieb$^{1}$, Angelica I Aviles-Rivero$^{1}$
\\  \:
$^{1}$ University of Cambridge, United Kingdom 
\\ $^{2}$ Univ. Bordeaux, CNRS, Bordeaux INP, IMB, UMR 5251, F-33400 Talence, France
\\ $^{3}$ The Hong Kong University of Science and Technology (Guangzhou), China
\\$^{4}$ The Hong Kong University of Science and Technology, HK SAR, China
}

\maketitle

\input{section/abstract}

\input{section/introduction}

\input{section/related_work}

\input{section/method}

\input{section/experiment}

\input{section/conclusion}

{\small
\bibliographystyle{ieee_fullname}
\bibliography{egbib}
}

\end{document}

%% file: section/abstract.tex
\begin{abstract}
Shadows in videos are difficult to detect because of the large shadow deformation between frames. In this work, we argue that accounting for shadow deformation is essential when designing a video shadow detection method. To this end, we introduce the shadow deformation attention trajectory (\soda), a new type of video self-attention module, specially designed to handle the large shadow deformations in videos. Moreover, we present a new shadow contrastive learning mechanism (\scotch) which aims at guiding the network to learn a unified shadow representation from massive positive shadow pairs across different videos. We demonstrate empirically the effectiveness of our two contributions in an ablation study. Furthermore, we show that \scotch~and \soda~significantly outperforms existing techniques for video shadow detection. 
Code is available at the project page: \url{https://lihaoliu-cambridge.github.io/scotch_and_soda/}

\vspace{-3mm}

\end{abstract}

%% file: section/introduction.tex
\section{Introduction} \label{sec:intro}

Shadow is an inherent part of videos, and they have an adverse effect on a wide variety of video vision tasks. Therefore, the development of robust video {shadow detection} techniques, to alleviate those negative effects, is of great interest for the community. Video shadow detection is usually formulated as a segmentation problem for videos, however and due to the nature of the problem, shadow detection greatly differs from other segmentation tasks such as object segmentation. For inferring the presence of shadows in an image, one has to account for the global content information such as light source orientation, and the presence of objects casting shadows. Importantly, in a given video, shadows considerably change appearance (deformation) from frame to frame due to light variation and object motion. Finally, shadows can span over different backgrounds over different frames, making approaches relying on texture information unreliable.

\begin{figure}[t!]
    \centering
    \includegraphics[width=\linewidth]{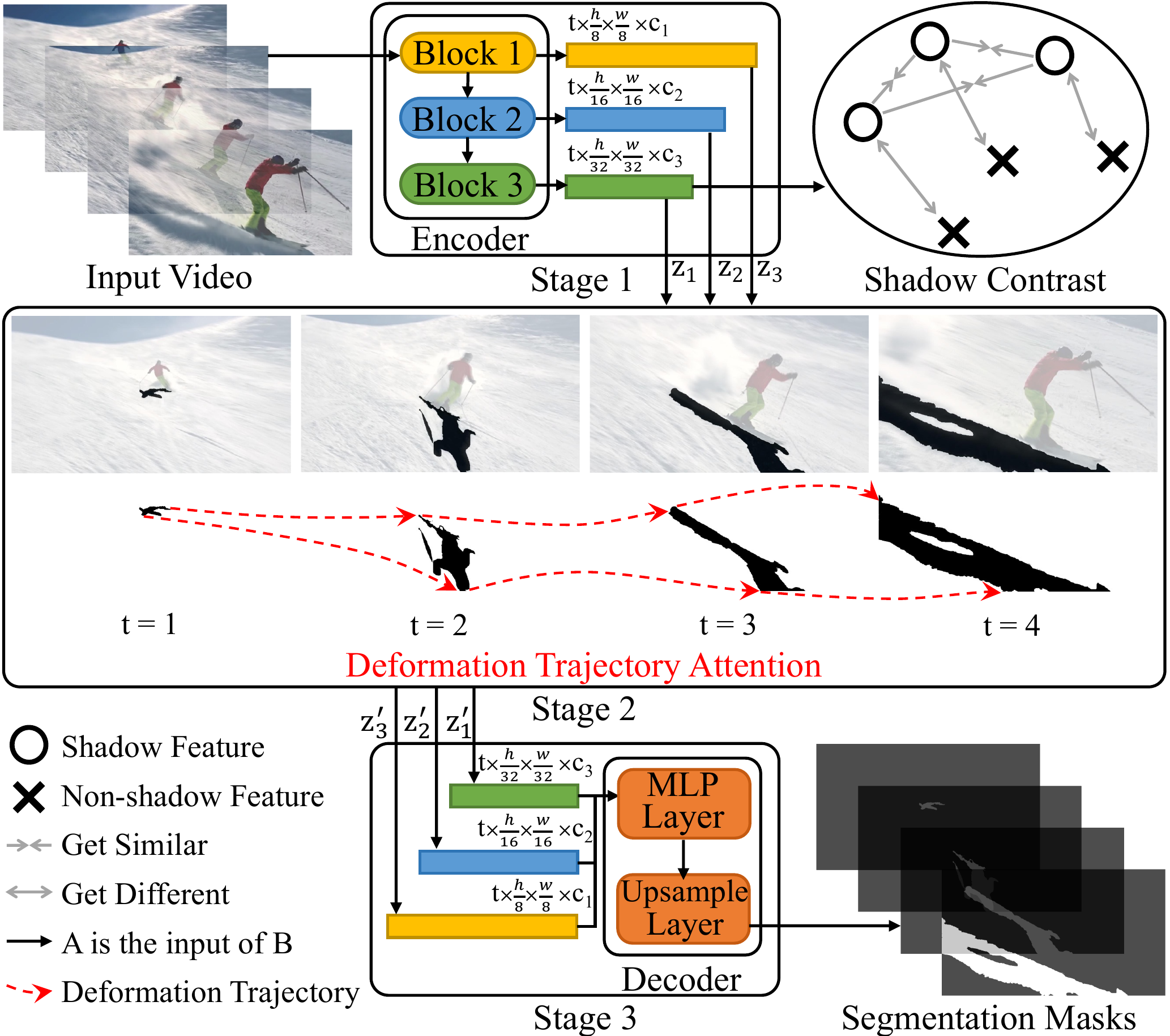}
    \caption{Overview of our \scotch~and \soda~framework. A MiT encoder extracts multi-scale features for each frame of the video (stage 1). Then, our deformation attention trajectory is applied to features individually to incorporate temporal information (stage 2). Finally, an MLP layer combines the multi-scale information to generate the segmentation masks (stage 3). The model is trained to contrast shadow and non-shadow features, by minimising our shadow contrastive loss with massive positive shadow pairs.}
    \label{fig:network_architecutre}
    \vspace{-3mm}
\end{figure}

Particularly, video shadow detection methods can be broadly divided into two main categories.
The first category refers to image shadow detection (ISD)~\cite{hu2019direction, zhu2018bidirectional, wang2020instance, wang2021single, zheng2019distraction, ding2022learning}. This family of techniques computes the shadow detection frame by frame. Although computationally saving, these methods are incapable of handling temporal information.
The second category refers to video shadow detection (VSD)~\cite{hu2021revisiting, chen2021triple, hu2021temporal, lu2022video, ding2022learning}. These methods offer higher performance as the analysis involves spatial-temporal information. Hence, our main focus is video shadow detection.

State-of-the-art video shadow detection methods rely on deep neural networks, which are trained on large annotated datasets. Specifically, those methods are composed of three parts: (i) a feature extraction network that extracts spatial features for each frame of the video: (ii) a temporal aggregation mechanism~\cite{chen2021triple, hu2021temporal} enriching spatial features with information from different frames; and (iii) a decoder, that maps video features to segmentation masks. Additionally, some works enforce consistency between frames prediction by using additional training criterion~\cite{lu2022video, ding2022learning}. We retain from these studies that the design of the temporal aggregation mechanism and the temporal consistency loss is crucial to the performance of a video shadow detection network, and we will investigate both of those aspects in this work. 

The current temporal aggregation mechanisms available in the literature were typically designed for video tasks such as video action recognition, or video object segmentation. Currently, the most widely used temporal aggregation mechanism is based on a variant of the self-attention mechanism~\cite{vaswani2017attention, arnab2021vivit, yan2022multiview, zhang2021vidtr, patrick2021keeping}. Recently, trajectory attention~\cite{patrick2021keeping} has been shown to provide state-of-the-art results on video processing.
Intuitively, trajectory attention aggregates information along the object's moving trajectory, while ignoring the context information, deemed as irrelevant. However, shadows in videos are subject to strong deformations, making them difficult to track, and thus they might cause the trajectory attention to fail.

In this work, we first introduce the ShadOw Deformation Attention trajectory (\soda), a spatial-temporal aggregation mechanism designed to better handle the large shadow deformations that occur in videos. \soda~operates in two steps. First, for each spatial location, an associated token is computed between the given spatial location and the video, which contains information in every time-step for the given spatial location. Second, by aggregating every associated spatial token, a new token is yielded with enriched spatial deformation information. Aggregating spatial-location-to-video information along the spatial dimension helps the network to detect shape changes in videos.

Besides, we introduce the Shadow COnTrastive meCHanism (\scotch), a supervised contrastive loss with massive positive shadow pairs aiming to drive our network to learn more discriminative features for the shadow regions in different videos. 
Specifically, in training, we add a contrastive loss at the coarsest layer of the encoder, driving the features from shadow regions close together, and far from the features from the non-shadow region. Intuitively, this contrastive mechanism drives the encoder to learn high-level representations of shadow, invariant to all the various factors of shadow variations, such as shape and illumination. 

In summary, our contributions are as follows:
\begin{itemize} [noitemsep,nolistsep]
    \item We introduce a new video shadow detection framework, in which we highlight:
    \begin{itemize}
        \item \soda, a new type of trajectory attention that harmonise the features of the different video frames at each resolution.
    
        \item \scotch, a contrastive loss that highlights a massive positive shadow pairs strategy in order to make our encoder learn more robust high-level representations of shadows.
         
    \end{itemize}
    \item We evaluate our proposed framework on the video shadow benchmark dataset ViSha~\cite{chen2021triple}, and compare with the state-of-the-art methods. Numerical and visual experimental results demonstrate that our approach outperforms, by a large margin, existing ones on video shadow detection. Furthermore, we provide an ablation study to further support the effectiveness of the technical contributions.
\end{itemize}

%% file: section/related_work.tex
\section{Related Work}\label{sec:related_work}

The task of {video} shadow detection has been extensively investigated in the community, in which solutions largely rely on analysing single frames (image shadow detection) or continuous multiple frames (video shadow detection). In this section, we review the existing techniques in turn, and then summarize the recent achievements in the video processing area to better illustrate the difference between existing work and our work.

\subsection{Image Shadow Detection}

Image shadow detection (ISD) can be cast as a semantic segmentation problem, where image object segmentation (IOS) methods can be used to solve this problem~\cite{lin2017feature,zhao2017pyramid,hou2017deeply,deng2018r3net,liu2022simultaneous,liu2020psi}.
However, IOS methods are not specifically designed for shadow detection. Hence, when re-training these methods directly for shadow detection, the performance is unsatisfactory due to the data bias. 

Techniques focused on image shadow detection incorporate problem-specific shadow knowledge into the model architecture and the training criterion ~\cite{hu2019direction, zhu2018bidirectional, wang2020instance, wang2021single, zheng2019distraction,chen2020multi,hu2021revisiting, ding2022learning}.
For example, BDRAR \cite{zhu2018bidirectional} introduces a bidirectional pyramidal architecture for shadow detection. 
DSD~\cite{zheng2019distraction} presents a distraction-aware shadow-module to reduce false positives.
Chen \textit{et al.}~\cite{chen2020multi} make use of non-labelled data, during training, with a task-specific semi-supervised learning mechanism called MTMT.  
Wang \textit{et al.} investigate the detection of shadows along with their corresponding objects~\cite{wang2020instance, wang2021single}.
Finally, FSDNet~\cite{hu2021revisiting} proposes a compact image shadow detection network. 
Whilst ISD techniques have demonstrated potential results, their performance on videos is limited by the lack of temporal information.

\subsection{Video Shadow Detection}

Another body of researchers has explored the task of shadow detection from the lens of video analysis. 
The work of~\cite{chen2021triple} proposes the TVSD model.
It relies on a dual-gated co-attention module, to aggregate features from different frames, and uses a contrastive learning mechanism to drive the encoder to discriminate frames from different videos.
Hu~\textit{et al.}~\cite{hu2021temporal} introduce an optical flow warping module to aggregate features from different frames. STICT~\cite{lu2022video} uses transfer learning, to transfer the knowledge of a supervised image shadow detection network to a video shadow detection network, without labelled videos, by training the network prediction to be consistent with respect to temporal interpolation~\cite{verma2022interpolation}. 
Moreover, the technique called SC-Cor~\cite{ding2022learning} presents a weakly supervised correspondence learning mechanism to enhance the temporal similarity of features corresponding to shadow region across frames. 

\subsection{Progresses in Video Processing}

The specific nature of videos, containing spatial and temporal information, has motivated the design of deep neural network architectures for different video processing applications. Models based on 3D CNN~\cite{tran2015learning, ji20123d} process videos by sequentially aggregating the spatio-temporal local information using 3D convolutional filters, but fail to effectively capture long-range temporal dependencies.
To alleviate this limitation, architectures using recurrent networks were introduced in~\cite{ballas2015delving, song2018pyramid}. Moreover, another set of works uses spatio-temporal memory bank mechanism~\cite{oh2019video} or spatio-temporal attention mechanism~\cite{vaswani2017attention} into the coarse layers of 3D CNN architectures ~\cite{wang2018non, lu2019see, li2019motion} to better integrate the temporal information for video processing. 

The success of the transformer network architecture, on a wide variety of vision tasks~\cite{dosovitskiy2020image,bulat2021space}, has motivated the use of transformer for video tasks. While the self-attention mechanism in transformers appears to be well suited to capture the long-range dependencies in videos, applying transformers to videos raises many challenges, such as the quadratic complexity in the input sequence length, and the large data requirement induced by the lack of problem-specific inductive bias. The works of~\cite{zhang2021vidtr, arnab2021vivit} propose to separate spatial and temporal attention to reduce computational complexity, and the authors of~\cite{yan2022multiview} propose to apply multiple encoders on multiple views of the video. 
Recently, trajectory attention~\cite{patrick2021keeping} was introduced as a way to incorporate an inductive bias in the self-attention operation to capture objects moving trajectories for better video recognition tasks.  

\subsection{Existing Works \& Comparison to Our Work}

All precedent works on VSD~\cite{chen2021triple, hu2021temporal, lu2022video, verma2022interpolation} rely on convolutional neural network architectures. \textit{To the best of our knowledge, our work is the first video shadow detection approach based on transformers.}

Moreover, whilst the work of~\cite{patrick2021keeping} also considers a type of attention trajectory,
their modelling hypothesis is that the video objects do not change shape over time. This is a strong assumption to fulfill for several vision applications such as shadow detection; as shadows significantly change shape from frame to frame. Our work first mitigates this issue by modelling in the trajectories the inherent deformation of the shadows.
Notice that TVSD~\cite{chen2021triple} uses an image-level contrastive loss between frames from different videos, and SC-Cor~\cite{ding2022learning} uses weakly-supervised correspondence learning for driving similar features from shadow-region close together. Unlike these works, we introduce a supervised contrastive strategy to contrast shadow features from non-shadow features. We underline that existing contrastive-based techniques assume that supervised contrastive learning is not performing better than its counterpart. In this work, we show that a well-designed supervised contrastive strategy indeed improves over existing works.

%% file: section/method.tex
\section{Methodology} 
\label{sec:method}
In this section, we introduce all the components of the video shadow detection method presented in this work. After a global description of our framework (i), we introduce \soda, our new video self-attention module (ii), and we describe the training criterion of our model, which includes \scotch, our shadow contrastive mechanism (iii).

\begin{figure*}[t!]
    \centering
    \includegraphics[width=\textwidth]{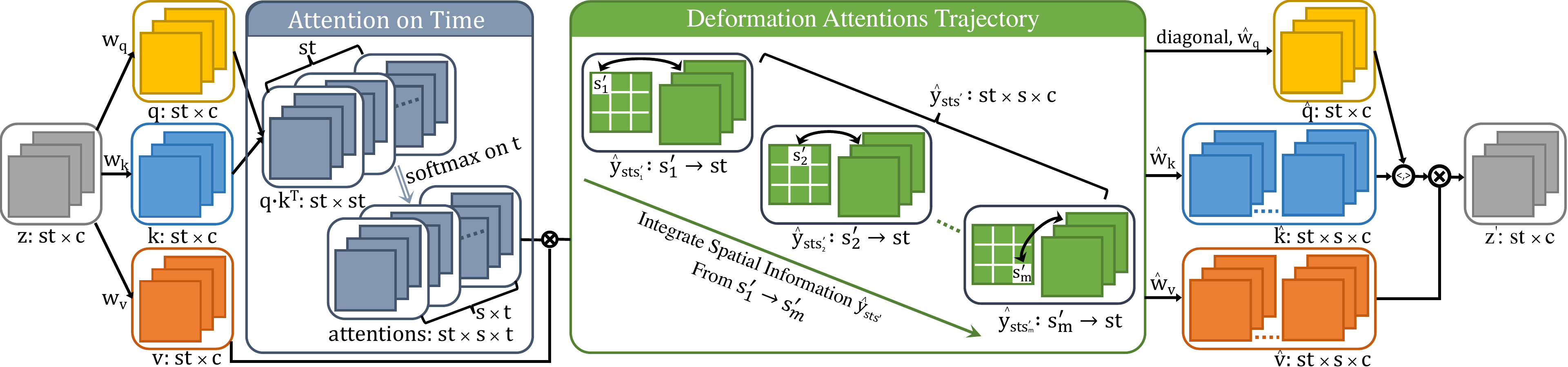}
    \caption{Deformation attention trajectory module. The input feature maps $z$ is used to generate $q$, $k$, $v$, respectively. The $q$ and $k$ are first used to calculate the pointwise $st$-to-$st$ similarity, followed by a softmax on $t$ to get the time attention. The time attention aggregated $v$ can generate $s$-to-$st$ (spatial-location-to-video) attention, named deformation attention. Then, the spatial-location-to-video attention is integrated along the spatial dimension to capture the deformation attention trajectory. Lastly, the deformation attention trajectory is used to generate the final feature map $z^{\prime}$ with a second self-attention ($\hat{q}$, $\hat{k}$, $\hat{v}$). In this figure, each square represents a spatial feature map. (The channel dimension is not represented for simplicity).
    }
    \label{fig:deformation_trajectory_attention}
    \vspace{-3mm}
\end{figure*}

\subsection{Framework Architecture Overview} \label{subsec:overall_framework}
Our proposed framework is composed of three main stages. The overall workflow is illustrated in Figure~\ref{fig:network_architecutre}, and details are provided next.

\smallskip
\faHandPointRight[regular]  \textbf{Stage 1: Feature Extraction.}
In this stage, Mix Transformer (MiT)~\cite{xie2021segformer} is adopted as the encoder. 
MiT takes video clips as input, and it outputs a set of different-resolution spatial feature maps. 
Unlike ViT~\cite{dosovitskiy2020image}, which only generates single-resolution feature maps, the hierarchical structure of MiT can generate CNN-like multi-level multi-scale feature maps.
These feature maps contain from high-resolution detailed information to low-resolution semantic information. By incorporating different levels of resolutions, the performance of tasks such as semantic segmentation can be boosted~\cite{xie2021segformer}.
In this stage, MiT only encodes the spatial information of the given video clips (input), that is, the temporal information is not yet incorporated. 

\faHandPointRight[regular]
\textbf{Stage 2: Harmonising Spatial and Temporal Information.}
The multi-scale feature maps from Stage 1 are then processed by the newly introduced shadow deformation attention trajectory module (\soda).
The goal of this module is to capture the shadow's deformation trajectory along the frames in the video clips.
Our deformation attention trajectory module processes independently each feature map with different-resolution. These processed feature maps are used as input to the decoder in the next stage.   

\faHandPointRight[regular] \textbf{Stage 3: Mask Shadow Generation. } 
In this stage, a light-weighted decoder, with only MLP layers, is adopted to reconstruct the shadow's segmentation masks. The decoder aims to mix the high- \& low- resolution information, from processed feature maps, for better semantic segmentation. The output of the decoder is a segmented video that marks out the shadows in each frame of the video.

\subsection{SODA: ShadOw Deformation Attention trajectory}
Transformers have revolutionised several tasks in the computer vision area. The key is the self-attention mechanism that can accommodate with any given type of data and domain.
However, for videos, the standard self-attention does not differentiate the spatial dimensions from the temporal dimension. This can lead the attention to focus on the redundant spatial information while neglecting the informative temporal variations in the videos. Nonetheless, video analysis is inherent to such temporality.
Most recently, trajectory attention~\cite{patrick2021keeping} has proposed to accommodate somehow with such issues.  However and even though trajectory attention has demonstrated potential results, it has a major limitation -- \textit{the objects are assumed not to change over time}. This is a major constraint in several video tasks including shadow detection; as shadows significantly undergo deformation from one frame to another. In this subsection, we introduce a new scheme called \textbf{S}had\textbf{O}w \textbf{D}eformation \textbf{A}ttention trajectory (\soda) to mitigate current drawbacks of the literature. 

Like in the classical self-attention setting, it begins with an input feature map generated from the encoder.
Specifically, let us denote $f_d\in\mathbb{R}^{t\times \frac{h}{d}\times \frac{w}{d}\times c}$, the generated feature map from the encoder, where $d$ is the spatial downsampling ratio, $c$ is the number of feature channels, and $t$ is the number of time frames in the video. $f_d$ is reshaped to a sequence of 1D token embedding denoted as $z\in\mathbb{R}^{n\times c}$, where $n=t\times \frac{h}{d}\times \frac{w}{d}$.
As shown in the left part of Figure~\ref{fig:deformation_trajectory_attention}, $z$ is then mapped to a set of query-key-value vectors $q, k, v\in\mathbb{R}^{n\times c}$ using linear projections $q=w_q\cdot z$, $k=w_k\cdot z$, $v=w_v\cdot z$, with projection matrices $w_q, w_k, w_v\in\mathbb{R}^{n\times n}$.

Our scheme considers two main parts: (i) temporal attention between the spatial location and video (Attention on Time in Figure~\ref{fig:deformation_trajectory_attention}), (ii) intra-space attention to capture deformation statues within a spatial scene (Deformation Attention Trajectory in Figure~\ref{fig:deformation_trajectory_attention}).
For the first part, given a space-time position in the video $st \in \{1, \cdots, n\}$, and a spatial location $s' \in \{1, \cdots, m\}$, where $m = \frac{n}{t}$,
the \textit{temporal attention} (deformation) between the space-time position $st$ and the spatial locations $s'$ is computed as:

\input{equation/equation_7}
\subsection{SCOTCH: Shadow COnTrastive meCHanism}
Contrastive learning~\cite{chen2020simple,liu2020contrastive,liu2022pc} has been proven to be an effective mechanism for learning distinctive features. By contrasting the positive pairs with high similarity and negative pairs with low similarity, the learned feature maps can be more discriminative in downstream tasks including classification and segmentation.
In previous video shadow detection task~\cite{chen2021triple}, positive and negative pairs are sampled from frames from the same video and from two different videos respectively.
Since the frames from one video have high similarity image content, the contrastive mechanism can help to discriminate different video content.

However, the key element in video shadow detection is the shadow itself instead of the video content. 
With the goal of boosting the detection performance, we introduce \scotch, a \textbf{S}hadow \textbf{CO}n\textbf{T}rastive me\textbf{CH}anism. \scotch~seeks to better guide the segmentation process for shadows and non-shadows regions in the videos.
Specifically, to learn a unified shadow feature for different videos, positive pairs are sampled from the shadow regions from different frames in different videos, whilst negative pairs are sampled as shadow and non-shadow regions on the frames in different videos.
\textit{We underline that unlike the classical contrastive loss used for unsupervised learning~\cite{oord2018representation,chen2020simple}, where there is only a small number of positive pairs, we proposed a massive positive shadow paired contrastive loss.} The key idea behind our loss is that -- we seek to not only maximise the difference between shadow and non-shadow features, but also maximise the similarity between features of shadows presented in different videos.
All shadow and non-shadow features are cropped from the last layer of the encoder presented in Section~\ref{subsec:overall_framework}, with the supervision of the segmentation masks.
The contrastive loss reads:
\input{equation/equation_10}

\textbf{Optimisation Scheme for Shadow Detection.}
Finally, to compute the shadow segmentation loss, we follow the default setting in~\cite{chen2021triple}. We use the binary cross entropy (BCE) loss with a lovasz-hinge loss~\cite{berman2018lovasz}. These two terms are added to define  the shadow segmentation loss as follows:
\input{equation/equation_11}

\noindent Our optimisation scheme is then given by \eqref{eq:eq10_contrastive} and \eqref{eq:eq11_seg}  as:
\input{equation/equation_12}

\input{table/table_1}

%% file: equation/equation_7.tex
\begin{equation}
\begin{aligned} \label{eq:eq7_defo_attention_1}
    \hat{y}_{s t s^{\prime}}=\sum_{t^{\prime}} v_{s^{\prime} t^{\prime}} \cdot \frac{\exp \langle q_{s t}, k_{s^{\prime} t^{\prime}} \rangle}{\sum_{\bar{t}} \exp \langle q_{s t}, k_{s^{\prime} \bar{t}} \rangle}
\end{aligned}
\end{equation}
For brevity, the notation is slightly abused by omitting the ``softmax operation on time dimension" in the  Fig.~\ref{fig:deformation_trajectory_attention} applied to the fraction, as well as the scaling parameter $\sqrt{n}$ (we will keep this notation convention throughout the paper).
The deformation encodes the connection between one space-time position and one spatial location, which indicates how the content of the space-time position $st$ is presented in spatial location $s'$. 

Once the temporal attentions are computed, the intra-space attention is then estimated to aggregate the spatial-location-to-video responses to space-level deformation. To do this, the computed deformation tokens are projected
to a new set of query-key-value vectors using linear projections:
\input{equation/equation_89}

%% file: equation/equation_89.tex
\begin{equation}
\begin{aligned} \label{eq:eq8_defo_attention_new_qkv}
    \hat{q}_{s t}=\hat{w}_{q} \cdot \hat{y}_{s t s}, \; \hat{k}_{s t s^{\prime}}=\hat{w}_{k} \cdot \hat{y}_{s t s^{\prime}}, \; \hat{v}_{s t s^{\prime}}=\hat{w}_{v} \cdot \hat{y}_{s t s^{\prime}}
\end{aligned}
\end{equation}
where $\hat{y}_{s t s}$ is the temporal connection from $st$ to the same spatial location $s$, and $\hat{q}_{s t}$ corresponds to the deformation reference point $st$ that is used to aggregate the location-to-video connection:
\begin{equation}
\begin{aligned} \label{eq:eq9_defo_attention_part2}
    \hat{y}_{s t}=\sum_{s^{\prime}} \hat{v}_{s t s^{\prime}} \cdot \frac{\exp \langle \hat{q}_{s t}, \hat{k}_{s t s^{\prime}} \rangle}{\sum_{\bar{s}} \exp \langle \hat{q}_{s t}, \hat{k}_{s t \bar{s}} \rangle}
\end{aligned}
\end{equation}
where $\hat{y}_{st}$ is the deformation attention output. The meaningful location-to-video tokens are pooled out to form the full space-level deformation status. By computing the intra-space attention, the attended feature map can capture the deformation status in different frames of the video, thus boosting the video shadow detection performance.

%% file: equation/equation_10.tex
\begin{equation}
\begin{split} \label{eq:eq10_contrastive}
    &\ell_{contrast}\left(\boldsymbol{v}, \boldsymbol{v}^{+}, \boldsymbol{v}^{-}\right)= \\ & -\log \left[\frac{\sum_{n=1}^{N} \exp \left(\boldsymbol{v} \cdot \boldsymbol{v}_{n}^{+} / \tau\right)}{\sum_{n=1}^{N} \exp \left(\boldsymbol{v} \cdot \boldsymbol{v}_{n}^{+} / \tau\right)+\sum_{n=1}^{N} \exp \left(\boldsymbol{v} \cdot \boldsymbol{v}_{n}^{-} / \tau\right)}\right]
\end{split} 
\end{equation}
where $\boldsymbol{v}\in\mathbb{R}^{c}$ is the query shadow feature. Moreover, $\boldsymbol{v}_{n}^{+}, \boldsymbol{v}_{n}^{-}\in\mathbb{R}^{n\times c}$ are the positive and negative groups respectively, and $\tau$ is a temperature hyperparameter. The final loss is computed across all frames in a mini-batch fashion.

%% file: equation/equation_11.tex
\begin{equation}
\begin{split} \label{eq:eq11_seg}
    \ell_{seg}=\ell_{bce} + \lambda_1 \ell_{hinge}
\end{split} 
\end{equation}

%% file: equation/equation_12.tex
\begin{equation}
\begin{split} \label{eq:eq12_all}
    \ell_{final}=\ell_{bce} + \lambda_1 \ell_{hinge} + \lambda_2 \ell_{contrast}
\end{split} 
\end{equation}
where $\lambda_1$ and $\lambda_2$ are two hyper-parameters weighting the relative effect of the hinge loss and the contrastive loss in the final loss. In the following experiments, $\lambda_1$, $\lambda_2$ were empirically set to a value of $1$ and $0.1$, respectively.

%% file: table/table_1.tex
\begin{table*}[ht]
\centering
\resizebox{0.70\textwidth}{!}{
\begin{tabular}{cc|cccc|cc}
\hline \toprule
\multicolumn{2}{c|}{\textsc{Methods}}                                                                                    & \multicolumn{6}{c}{\textsc{Evaluation Metrics}}                                                                       \\ \hline
\multicolumn{1}{c|}{Tasks}                & \multicolumn{1}{c|}{Techniques}                                       & \multicolumn{1}{p{1.3cm}}{\,\,\,MAE $\downarrow$} & \multicolumn{1}{p{1.3cm}}{\,\,\,\,\,$\textrm{F}_{\beta} \uparrow$}  & \multicolumn{1}{p{1.3cm}}{\,\,\,\,\,IoU $\uparrow$}   & \multicolumn{1}{p{1.3cm}|}{\,\,\,\,BER $\downarrow$}  & \multicolumn{1}{p{1.3cm}}{\,S-BER $\downarrow$} & \multicolumn{1}{p{1.3cm}}{N-BER $\downarrow$} \\ \midrule
\multicolumn{1}{c|}{\multirow{4}{*}{IOS}} & \multicolumn{1}{c|}{$\star$ FPN~\cite{lin2017feature}}              & 0.044   & 0.707   & 0.512   & 19.49     & 36.59  & 2.40         \\
\multicolumn{1}{c|}{}                     & \multicolumn{1}{c|}{PSPNet~\cite{zhao2017pyramid}}                  & 0.051   & 0.642   & 0.476   & 19.75                                          & 36.44  & 3.07         \\
\multicolumn{1}{c|}{}                     & \multicolumn{1}{c|}{DSS~\cite{hou2017deeply}}                       & 0.045   & 0.696   & 0.502   & 19.77                                          & 36.96  & 2.59         \\
\multicolumn{1}{c|}{}                     & \multicolumn{1}{c|}{$\textrm{R}^{3}$Net~\cite{deng2018r3net}}       & 0.044   & 0.710   & 0.502   & 20.40                                          & 37.37  & 3.55         \\ \midrule[0.8pt]
\multicolumn{1}{c|}{\multirow{4}{*}{ISD}} & \multicolumn{1}{c|}{BDRAR~\cite{zhu2018bidirectional}}              & 0.050   & 0.695   & 0.484   & 21.29                                          & 40.28  & 2.31         \\ 
\multicolumn{1}{c|}{}                     & \multicolumn{1}{c|}{$\star$ DSD~\cite{zheng2019distraction}}        & 0.043   & 0.702   & 0.518   & 19.88     & 37.89  & 1.88         \\ 
\multicolumn{1}{c|}{}                     & \multicolumn{1}{c|}{MTMT~\cite{chen2020multi}}                      & 0.043   & 0.729   & 0.517   & 20.28                                          & 38.71  & 1.86         \\ 
\multicolumn{1}{c|}{}                     & \multicolumn{1}{c|}{FSDNet~\cite{hu2021revisiting}}                 & 0.057   & 0.671   & 0.486   & 20.57                                          & 38.06  & 3.06         \\ \midrule
\multicolumn{1}{c|}{\multirow{5}{*}{VOS}} & \multicolumn{1}{c|}{PDBM~\cite{song2018pyramid}}                    & 0.066   & 0.623   & 0.466   & 19.73                                          & 34.32  & 5.16         \\
\multicolumn{1}{c|}{}                     & \multicolumn{1}{c|}{COSNet~\cite{lu2019see}}                        & 0.040   & 0.705   & 0.514   & 20.50                                          & 39.22  & 1.79         \\
\multicolumn{1}{c|}{}                     & \multicolumn{1}{c|}{$\star$ FEELVOS~\cite{voigtlaender2019feelvos}} & 0.043   & 0.710   & 0.512   & 19.76     & 37.27  & 2.26         \\
\multicolumn{1}{c|}{}                     & \multicolumn{1}{c|}{STM~\cite{oh2019video}}                         & 0.068   & 0.597   & 0.408   & 25.69                                          & 47.44  & 3.96         \\ \midrule
\multicolumn{1}{c|}{\multirow{4}{*}{VSD}} & \multicolumn{1}{c|}{TVSD~\cite{chen2021triple}}                     & 0.033   & 0.757   & 0.567   & 17.70                                          & 33.97  & 1.45         \\ 
\multicolumn{1}{c|}{}                     & \multicolumn{1}{c|}{STICT~\cite{lu2022video}}                       & 0.046   & 0.702   & 0.545   & 16.60                                          & 29.58  & 3.59         \\ 
\multicolumn{1}{c|}{}                     & \multicolumn{1}{c|}{SC-Cor~\cite{ding2022learning}}                 & 0.042   & 0.762   & 0.615   & 13.61                                          & 24.31  & 2.91         \\ \cmidrule{2-8} 
\multicolumn{1}{c|}{}                     & \multicolumn{1}{c|}{$\star$ \scotch~and \soda \enspace}                          & \textbf{0.029}   & \textbf{0.793}   & \textbf{0.640}   & \textbf{9.066}      & \textbf{16.26}  & \textbf{1.44}         \\ \bottomrule
\end{tabular}
}
\caption{Comparisons between our proposed technique and SOTA techniques on the ViSha dataset. ``MAE" denotes mean absolute error, ``$\textrm{F}_{\beta}$" denotes F-measure score, ``IoU" denotes intersection over union, ``BER" denotes balance error rate, and ``S-BER" means shadow BER, ``N-BER" means non-shadow BER. The $\uparrow$ denotes the higher the value is the better the performance is, whilst the $\downarrow$ means the opposite. $\star$ indicates the best performed network in each category.
}
\label{table_1_numerical_comparision}
\vspace{-3mm}
\end{table*}

%% file: section/experiment.tex
\section{Experimental Results}\label{sec:experiment}
This section details all experiments performed to validate our proposed framework.

\subsection{Dataset and Evaluation Metrics}

\textbf{Data Description.} 
We utilise the largest and latest \textbf{Vi}deo \textbf{Sha}dow dataset (\textbf{ViSha})~\cite{chen2021triple} to evaluate the effectiveness of our proposed VSD method. 
The ViSha dataset has 120 videos, and each video contains between 29 and 101 frames. ViSha is composed of a total of 11,685 frames corresponding to a total duration of 390 seconds of video.

\textbf{Data Pre-processing.} 
We follow the setting introduced in ViSha~\cite{chen2021triple}.
That is, we use the same train-test split, with 50 videos for training and 70 videos for testing. 
During training, we also use the same data augmentation strategy as ~\cite{chen2021triple} to enrich the variety of the dataset. Specifically, during training, images are re-scaled to size $512\times 512$, and are randomly flipped horizontally. In testing, only re-scaling to the unified size $512\times 512$ is used. 

\textbf{Evaluation Metrics}. Following the evaluation protocol used in~\cite{chen2021triple, lu2022video, ding2022learning}, we employ four common evaluation metrics to measure the shadow detection accuracy: MAE, $\textrm{F}_{\beta}$, IoU, and BER.
Lower MAE and BER scores, and higher $\textrm{F}_{\beta}$ and IoU scores indicate a better video shadow detection result.
Moreover, we also provide the shadow BER (S-BER) and the non-shadow BER scores (N-BER) 
to further compare different VSD methods.

\textbf{Implementation Details.}
Our proposed segmentation architecture is built using the PyTorch-lightning~\cite{falcon2019pytorch} deep-learning framework. 
The parameters of the feature extraction encoder are initialised using the weights from the MiT-B3 model pre-trained for image segmentation on ADE20K dataset~\cite{zhou2017scene, zhou2019semantic}, publicly available on HuggingFace~\cite{wolf2019huggingface}. The remaining parameters (attention modules and the MLP decoder) are randomly initialised using ``Xavier" methods~\cite{glorot2010understanding}.
During training, AdamW optimizer~\cite{loshchilov2017decoupled} is used with an initial learning rate of $1\times 10^{-6}$ 
without decay.
All experiments and ablation studies are
trained for 36 epochs, for a training time of approximately 12 hours on NVIDIA A100 GPU with 80G RAM with batch size of 8. %

\begin{figure*}[t]
    \centering
    \includegraphics[width=\textwidth]{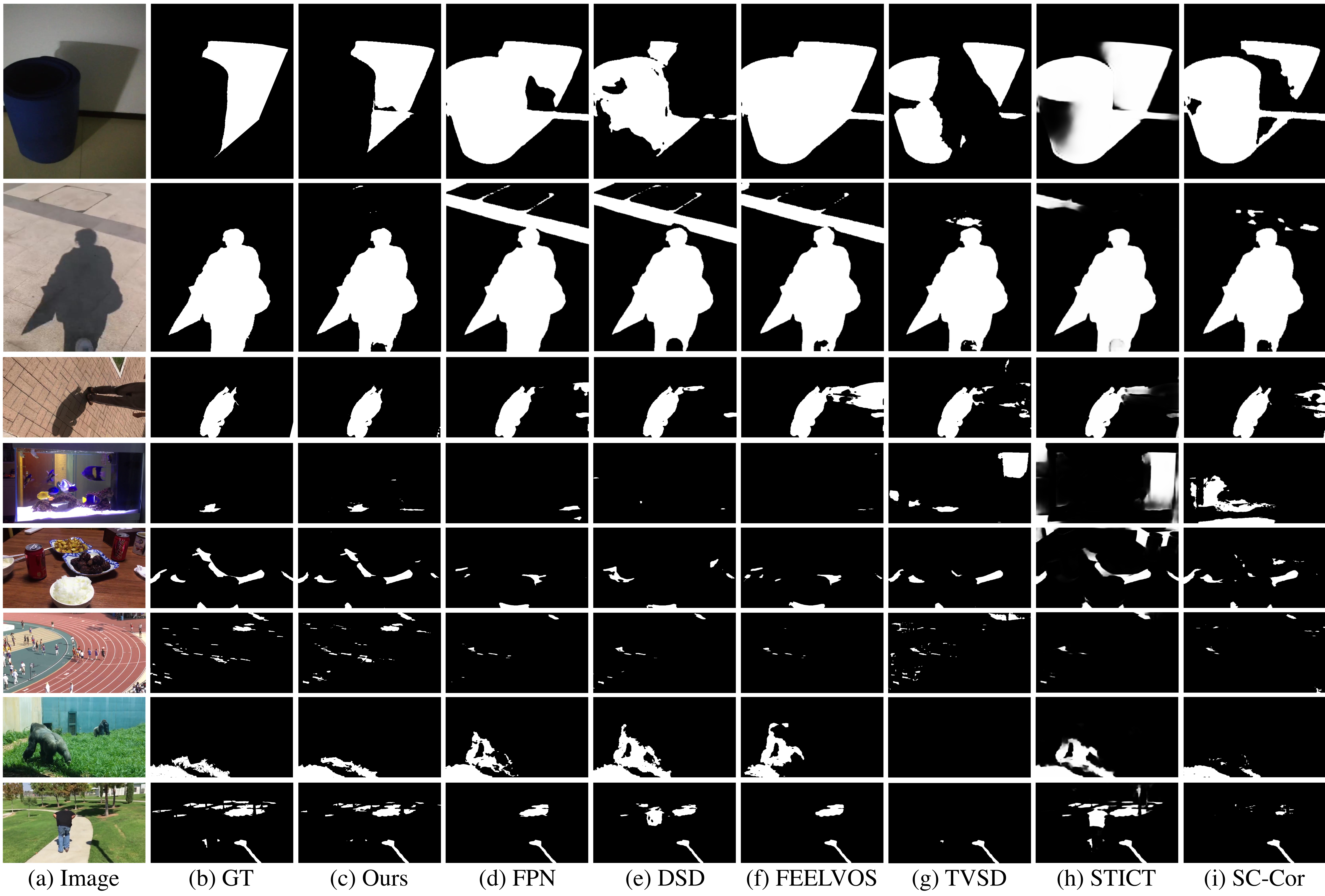}
    \caption{Visual comparisons of video shadow detection results produced by our network (\scotch~and~\soda) and compared methods. Apparently, our method can more accurately identify shadow pixels and our results are more consistent with the ground truth in the 2nd column. The video segmentation results can be found on the project page. (d-e) are the methods with the highest performance in IOS, ISD, and VOS in Table~\ref{table_1_numerical_comparision}, whilst (g-i) are all from the VSD area.}
    \label{fig:segmentation_results}
    \vspace{-3mm}
\end{figure*}

\subsection{Comparison to SOTA Techniques}

\textbf{Compared Methods.}
Video shadow detection is a relatively recent topic, and there are only three directly-related methods designed for this task. Hence, following existing VSD methods, we make comparisons against 4 different kinds of methods, including image object segmentation (IOS), image shadow detection (ISD), video object segmentation (VOS), and video shadow detection (VSD). 
For IOS, they are FPN~\cite{lin2017feature}, PSPNet~\cite{zhao2017pyramid}, DSS~\cite{hou2017deeply}, and $\textrm{R}^3$-Net~\cite{deng2018r3net}, while the ISD methods are BDRAR~\cite{zhu2018bidirectional}, DSD~\cite{zheng2019distraction}, MTMT~\cite{chen2020multi}, and FSDNet~\cite{hu2021revisiting}.
The compared VOS methods include PDBM~\cite{song2018pyramid}, COSNet~\cite{lu2019see},
FEELVOS~\cite{voigtlaender2019feelvos}, and STM~\cite{oh2019video}, while compared VSD methods are TVSD~\cite{chen2021triple}, STICT~\cite{lu2022video}, and SC-Cor~\cite{ding2022learning}. 
We obtain the results by re-training their network parameters with unified training parameters or by downloading the results from the TVSD~\cite{chen2021triple} repository.

\textbf{Quantitative Comparisons.}
Table~\ref{table_1_numerical_comparision} summarises MAE, $\textrm{F}_{\beta}$, IoU, and BER scores of our network against SOTA techniques.
For each category, we use $\star$ to mark out the method with the best performance. 
From these quantitative results, we observe that the IOS and ISD report readily competing results. These results are expected as the modelling hypothesis for both families of techniques relies on only the image level analysis. Whilst VOS techniques consider also temporal information, these techniques are customised as a general framework for video object segmentation. However, shadow detection is more complex due to the fast change in appearance between frames. Notably, we observe that VSD techniques indeed provide a substantial performance improvement compared to other methods; as they are designed considering the complexity of shadows. 

More importantly, our method outperforms all other techniques by a significant margin for all evaluation metrics -- further supporting the superiority of our approach. In particular, our method yields to a balanced error rate of $9.066$ which is more than $4$ points below SC-Cor, the latest SOTA technique, in terms of BER. The error rate for the shadow label (denoted as S-BER in Table~\ref{table_1_numerical_comparision}) of our method is 8 points below SC-Cor. We also underline that our significant improvement in performance comes with a negligible computational cost. We report the test time in Table~\ref{table_3_model_analysis}, where we observe that our \scotch~and \soda~framework only requires a fraction of time than compared methods.

\textbf{Visual Comparison.}
Figure~\ref{fig:segmentation_results} visually compares the shadow segmentation masks from our method and the compared methods on different input video frames.
For video frames with black objects at the first three rows, we find that compared methods tend to wrongly identify those black objects as shadow ones, while our method can still accurately detect shadow pixels under the corruption of black objects.
Moreover, compared methods tend to miss some shadow regions when the input video frames contain multiple shadow detection, as shown in the 4-th row to the 8-th row of Figure~\ref{fig:segmentation_results}. 
On the contrary, our method can identify all these shadow regions of input video frames, and our detection results are most consistent with the ground truth in the 2nd column of Figure~\ref{fig:segmentation_results}. We also provide the video segmentation masks on the project page to demonstrate the temporal coherence of the results provided by our method. 

\input{table/table_3}

\textbf{Model size and inference time.}
Table~\ref{table_3_model_analysis} further compares our network and three state-of-the-art video shadow detection methods in terms of the model size (Params), computational complexity (GMACs), inference time (Time), and segmentation accuracy (BER).
Apparently, among the three compared VSD methods, STICT has the smallest testing time (13.5 minutes).
Compared to STICT, our method further reduces the inference time from 13.5 minutes to 9.15 minutes to test all 70 testing videos with 6,897 images.
Although our method takes 2nd rank in terms of the model size and computational complexity, they are only larger than STICT. 
This is because we only use a light-weighted MLP layer as the decoder to integrate the multi-resolution feature maps, which is computational-saving.
In terms of performance, our method has a superior BER performance than STICT by reducing the BER score from 16.60 to 9.006, which indicates that with a minor compromise on the model size, our method can more accurately identify video shadows than STICT, TVSD, and SC-Cor.

\input{table/table_2}

\subsection{Ablation Study}
In Table~\ref{table_2_ablation_study}, we perform ablation studies on our main contributions to evaluate the effectiveness of each component.

\textbf{Baseline with MiT backbone.}
In order to evaluate the role of the MiT backbone on the final performance of our method, we define the segmentation network with MiT backbone as the baseline, which is trained by using the classical segmentation loss~\eqref{eq:eq11_seg}, but \emph{without} the deformation attention trajectory and shadow level contrastive mechanism. 
This baseline already provides results on par or better than the previous three works in the VSD area (see Table~\ref{table_1_numerical_comparision} for comparison), even without considering any kind of temporal information. This illustrates the superior performance of the MiT transformer-based architecture over convolutional architectures on the task of video shadow detection.

\textbf{Attention mechanisms.} 
We then evaluate the effectiveness of different attention mechanisms, including trajectory attention~\cite{patrick2021keeping} and our newly introduced shadow deformation attention. Both types of attention modules provide an improvement over the baseline, which was to be expected as those modules give the ability to consider the temporal information within the videos.
Our deformation attention trajectory module also appears to provide better results than the trajectory attention, indicating the importance of considering the deformation in the design of the attention module. %

\textbf{Contrastive losses.} 
Next, we compare the effect of two types of contrastive criterion, the image level contrastive loss used in TVSD \cite{chen2021triple}, and our feature-level shadow contrastive loss. 
The inter-frame contrastive learning slightly improves the baseline, whereas our shadow contrastive loss provides a significant improvement over the baseline. Those results illustrate the superiority of the features-level contrastive loss over the frame-level contrastive loss. 

\textbf{Final model.} 
\scotch~reduces the variance of the spatial shadow features, while \soda~processes the spatial features at different time-step to consider the temporal information in the video. Thus, \scotch~and \soda~have a complementary effect, as feeding \soda~with more robust spatial features from \scotch, we are able to reach the best performance, outperforming all the previous settings.

\begin{figure}[t]
    \centering
    \includegraphics[width=\linewidth]{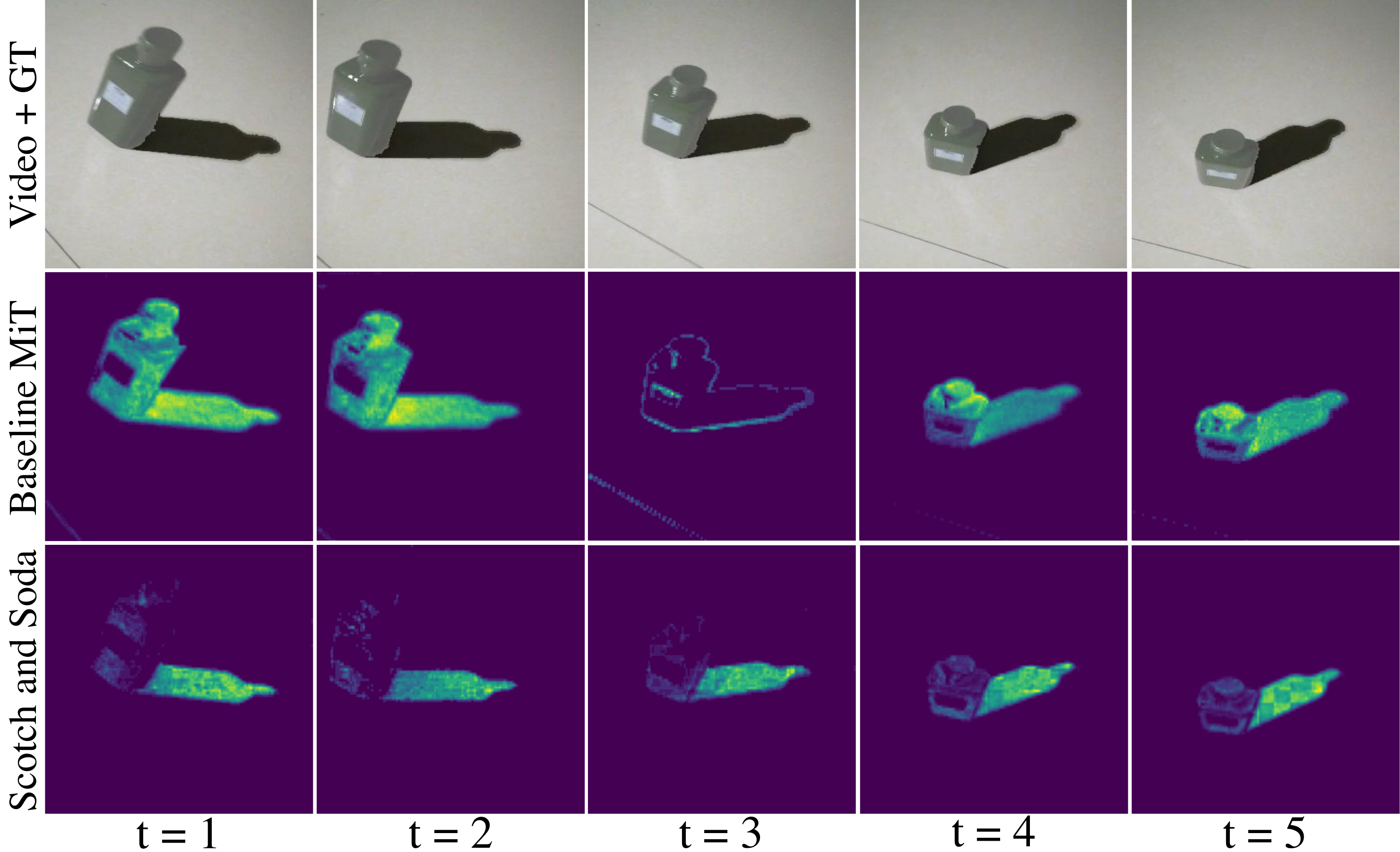}
    \caption{Attention maps visualisation. The low-resolution layers of the MiT encoder are selected for the visualization, with the query from the center of the ground truth shadow mask. From top to bottom row are the input video blended with segmentation mask, attention maps for baseline MiT methods, and attention maps from ours \scotch~and~\soda~with deformation attention trajectory and shadow contrastive mechanism. 
    }
    \label{fig:attention_map}  
    \vspace{-3mm}
\end{figure}

\subsection{Attention Map Visualisation}
In Figure~\ref{fig:attention_map}, we visually compare the attention maps from the baseline model (second row) with our proposed \scotch~and~\soda~(third row). 
We can observe that the baseline model highlights the object part, whilst ours focuses more on the shadow region with the help of contrastive learning. 
We can also observe that the baseline model might lose track of shape in different frames (third column), whilst our approach provides consistent shape information at different times with the help of the deformation attention trajectory.
Hence, the model using the shadow deformation attention trajectory and the shadow contrastive loss is better at tracking the shadow region during different frames, while ignoring the non-shadow part of the image.

%% file: table/table_3.tex
\begin{table}[t!]
\centering
\resizebox{\linewidth}{!}{
\begin{tabular}{c|c|c|c|c}
\hline \toprule
Networks                                    & \multicolumn{1}{p{1.3cm}|}{\,\,Params}             & \multicolumn{1}{p{1.3cm}|}{\,\,GMACs}                   & \multicolumn{1}{p{1.3cm}|}{\,\,\,\,\,Time}     &  \multicolumn{1}{p{1.3cm}}{\,\,\,\,\,\,BER}       \\ \midrule
TVSD~\cite{chen2021triple}                  & 243.32             & 158.89                     & 32.4                & 17.70             \\ 
STCIT~\cite{lu2022video}                    & \textbf{104.68}    & \textbf{40.99}             & \underline{13.5}    & 16.60             \\ 
SC-Cor~\cite{ding2022learning}              & 232.63             & 218.4                      & 21.8                & \underline{13.61} \\ \midrule
\scotch~and~\soda                                        & \underline{211.79} & \underline{122.46}         & \textbf{9.15}       & \textbf{9.006}     \\ \bottomrule
\end{tabular}
}
\caption{Comparison of model size (Params), computational complexity (GMACs), inference time (Time), and segmentation accuracy (BER). Specifically, the units for Params and Time are (MB) and (Mins), respectively. We denote the \textbf{best} and \underline{second best} in \textbf{bold} and \underline{underline} font.}
\label{table_3_model_analysis}
\vspace{-3mm}
\end{table}

%% file: table/table_2.tex
\begin{table}[t]
\centering
\resizebox{\linewidth}{!}{
\begin{tabular}{ccc|cccc}
\hline \toprule
\multicolumn{3}{c|}{Components}                                                                                              & \multicolumn{4}{c}{Evaluation Metrics}                                       \\ \hline
\multicolumn{1}{c|}{Backbone}            & \multicolumn{1}{c|}{Attention}     & \multicolumn{1}{c|}{Contrast}              & \multicolumn{1}{c}{MAE $\downarrow$} & \multicolumn{1}{c}{$\textrm{F}_{\beta}$ $\uparrow$} & \multicolumn{1}{c}{IoU $\uparrow$}   & \multicolumn{1}{c}{BER $\downarrow$}   \\ \midrule
\multicolumn{1}{c|}{MiT}                 & \multicolumn{1}{c|}{\xmark}         & \multicolumn{1}{c|}{\xmark}                                 & 0.048   & 0.755   & 0.584   & 13.18          \\ \midrule
\multicolumn{1}{c|}{MiT}                 & \multicolumn{1}{c|}{Trajectory}     & \multicolumn{1}{c|}{\xmark}                                 & 0.048   & 0.760   & 0.593   & 12.35          \\
\multicolumn{1}{c|}{MiT}                 & \multicolumn{1}{c|}{\soda}          & \multicolumn{1}{c|}{\xmark}                & \dashuline{0.041}   & \underline{0.791}   & \underline{0.613}   & \underline{10.55}          \\ \midrule
\multicolumn{1}{c|}{MiT}                 & \multicolumn{1}{c|}{\xmark}         & \multicolumn{1}{c|}{Image}                                  & 0.048   & 0.761   & 0.588   & 12.85          \\ 
\multicolumn{1}{c|}{MiT}                 & \multicolumn{1}{c|}{\xmark}         & \multicolumn{1}{c|}{Shadow}                                  & \dashuline{0.041}   & 0.767   & 0.592   & 12.02          \\ 
\multicolumn{1}{c|}{MiT}                 & \multicolumn{1}{c|}{\xmark}         & \multicolumn{1}{c|}{\scotch}                         & \underline{0.034}   & \dashuline{0.771}   & \dashuline{0.606}   & \dashuline{11.29}        \\ \midrule
\multicolumn{1}{c|}{\textbf{$\dagger$} MiT \enspace}         & \multicolumn{1}{c|}{\soda}            & \multicolumn{1}{c|}{\scotch}       & \textbf{0.029}   & \textbf{0.793}   & \textbf{0.640}   & \textbf{9.066}          \\ \bottomrule
\end{tabular}
}
\caption{Ablation study on different components of our proposed methods on the ViSha dataset. The $\uparrow$ denotes the higher the value is the better the performance is, whilst the $\downarrow$ means the opposite. ``$\dagger$" denotes our final methods with the highest performance in all evaluation metrics. Notations : \textbf{best}, \underline{second best}, \dashuline{third best}.
}
\label{table_2_ablation_study}
\vspace{-3mm}
\end{table}

%% file: section/conclusion.tex
\section{Conclusion}
\label{sec:conclusion}

In this paper, we introduced \scotch~and \soda, a new transformer video shadow detection framework. We developed shadow deformation attention trajectory (\soda), a self-attention module specially designed to handle the shadow deformation in videos, and we introduced a shadow contrastive mechanism (\scotch) to guide our network to better discriminate between shadow and non-shadow features. We demonstrate the effectiveness of the contributions with ablation studies. Finally, we show that our proposed method outperforms by a large margin concurrent video shadow segmentation works on the ViSha dataset. 

\vspace{2mm}
\noindent \textbf{Acknowledgements.}
This work is supported by Girton Postgraduate Research Scholarships, GSK Ph.D. Scholarship, CMIH, CCIMI, Philip Leverhulme Prize, Royal Society Wolfson Fellowship, EPSRC Advanced Career Fellowship EP/V029428/1, EPSRC grants EP/S026045/1, EP/T003553/1, EP/N014588/1, EP/T017961/1, Wellcome Innovator Awards 215733/Z/19/Z and 221633/Z/20/Z, EU Horizon 2020 research and innovation programme under the Marie Skodowska-Curie grant agreement No. 777826 NoMADS, and Guangzhou Municipal Science and Technology Project Grant No. 2023A03J0671.

%% file: PaperForReview.bbl
\begin{thebibliography}{10}\itemsep=-1pt

\bibitem{arnab2021vivit}
Anurag Arnab, Mostafa Dehghani, Georg Heigold, Chen Sun, Mario Lu{\v{c}}i{\'c},
  and Cordelia Schmid.
\newblock Vivit: A video vision transformer.
\newblock In {\em Proceedings of the IEEE/CVF International Conference on
  Computer Vision}, pages 6836--6846, 2021.

\bibitem{ballas2015delving}
Nicolas Ballas, Li Yao, Chris Pal, and Aaron Courville.
\newblock Delving deeper into convolutional networks for learning video
  representations.
\newblock {\em arXiv preprint arXiv:1511.06432}, 2015.

\bibitem{berman2018lovasz}
Maxim Berman, Amal~Rannen Triki, and Matthew~B Blaschko.
\newblock The lov{\'a}sz-softmax loss: A tractable surrogate for the
  optimization of the intersection-over-union measure in neural networks.
\newblock In {\em Proceedings of the IEEE conference on computer vision and
  pattern recognition}, pages 4413--4421, 2018.

\bibitem{bulat2021space}
Adrian Bulat, Juan~Manuel Perez~Rua, Swathikiran Sudhakaran, Brais Martinez,
  and Georgios Tzimiropoulos.
\newblock Space-time mixing attention for video transformer.
\newblock {\em Advances in Neural Information Processing Systems},
  34:19594--19607, 2021.

\bibitem{chen2020simple}
Ting Chen, Simon Kornblith, Mohammad Norouzi, and Geoffrey Hinton.
\newblock A simple framework for contrastive learning of visual
  representations.
\newblock In {\em International conference on machine learning}, pages
  1597--1607. PMLR, 2020.

\bibitem{chen2021triple}
Zhihao Chen, Liang Wan, Lei Zhu, Jia Shen, Huazhu Fu, Wennan Liu, and Jing Qin.
\newblock Triple-cooperative video shadow detection.
\newblock In {\em Proceedings of the IEEE/CVF Conference on Computer Vision and
  Pattern Recognition}, pages 2715--2724, 2021.

\bibitem{chen2020multi}
Zhihao Chen, Lei Zhu, Liang Wan, Song Wang, Wei Feng, and Pheng-Ann Heng.
\newblock A multi-task mean teacher for semi-supervised shadow detection.
\newblock In {\em Proceedings of the IEEE/CVF Conference on computer vision and
  pattern recognition}, pages 5611--5620, 2020.

\bibitem{deng2018r3net}
Zijun Deng, Xiaowei Hu, Lei Zhu, Xuemiao Xu, Jing Qin, Guoqiang Han, and
  Pheng-Ann Heng.
\newblock R3net: Recurrent residual refinement network for saliency detection.
\newblock In {\em Proceedings of the 27th International Joint Conference on
  Artificial Intelligence}, pages 684--690. AAAI Press Menlo Park, CA, USA,
  2018.

\bibitem{ding2022learning}
Xinpeng Ding, Jingweng Yang, Xiaowei Hu, and Xiaomeng Li.
\newblock Learning shadow correspondence for video shadow detection.
\newblock {\em arXiv preprint arXiv:2208.00150}, 2022.

\bibitem{dosovitskiy2020image}
Alexey Dosovitskiy, Lucas Beyer, Alexander Kolesnikov, Dirk Weissenborn,
  Xiaohua Zhai, Thomas Unterthiner, Mostafa Dehghani, Matthias Minderer, Georg
  Heigold, Sylvain Gelly, et~al.
\newblock An image is worth 16x16 words: Transformers for image recognition at
  scale.
\newblock {\em arXiv preprint arXiv:2010.11929}, 2020.

\bibitem{falcon2019pytorch}
William Falcon et~al.
\newblock Pytorch lightning.
\newblock {\em GitHub. Note: https://github.
  com/PyTorchLightning/pytorch-lightning}, 3(6), 2019.

\bibitem{glorot2010understanding}
Xavier Glorot and Yoshua Bengio.
\newblock Understanding the difficulty of training deep feedforward neural
  networks.
\newblock In {\em Proceedings of the thirteenth international conference on
  artificial intelligence and statistics}, pages 249--256. JMLR Workshop and
  Conference Proceedings, 2010.

\bibitem{hou2017deeply}
Qibin Hou, Ming-Ming Cheng, Xiaowei Hu, Ali Borji, Zhuowen Tu, and Philip~HS
  Torr.
\newblock Deeply supervised salient object detection with short connections.
\newblock In {\em Proceedings of the IEEE conference on computer vision and
  pattern recognition}, pages 3203--3212, 2017.

\bibitem{hu2021temporal}
Shilin Hu, Hieu Le, and Dimitris Samaras.
\newblock Temporal feature warping for video shadow detection.
\newblock {\em arXiv preprint arXiv:2107.14287}, 2021.

\bibitem{hu2019direction}
Xiaowei Hu, Chi-Wing Fu, Lei Zhu, Jing Qin, and Pheng-Ann Heng.
\newblock Direction-aware spatial context features for shadow detection and
  removal.
\newblock {\em IEEE transactions on pattern analysis and machine intelligence},
  42(11):2795--2808, 2019.

\bibitem{hu2021revisiting}
Xiaowei Hu, Tianyu Wang, Chi-Wing Fu, Yitong Jiang, Qiong Wang, and Pheng-Ann
  Heng.
\newblock Revisiting shadow detection: A new benchmark dataset for complex
  world.
\newblock {\em IEEE Transactions on Image Processing}, 30:1925--1934, 2021.

\bibitem{ji20123d}
Shuiwang Ji, Wei Xu, Ming Yang, and Kai Yu.
\newblock 3d convolutional neural networks for human action recognition.
\newblock {\em IEEE transactions on pattern analysis and machine intelligence},
  35(1):221--231, 2012.

\bibitem{li2019motion}
Haofeng Li, Guanqi Chen, Guanbin Li, and Yizhou Yu.
\newblock Motion guided attention for video salient object detection.
\newblock In {\em Proceedings of the IEEE/CVF international conference on
  computer vision}, pages 7274--7283, 2019.

\bibitem{lin2017feature}
Tsung-Yi Lin, Piotr Doll{\'a}r, Ross Girshick, Kaiming He, Bharath Hariharan,
  and Serge Belongie.
\newblock Feature pyramid networks for object detection.
\newblock In {\em Proceedings of the IEEE conference on computer vision and
  pattern recognition}, pages 2117--2125, 2017.

\bibitem{liu2020contrastive}
Lihao Liu, Angelica~I Avil{\'e}s-Rivero, and Carola-Bibiane Sch{\"o}nlieb.
\newblock Contrastive registration for unsupervised medical image segmentation.
\newblock {\em arXiv preprint arXiv:2011.08894}, 2020.

\bibitem{liu2022simultaneous}
Lihao Liu, Chenyang Hong, Angelica~I Aviles-Rivero, and Carola-Bibiane
  Sch{\"o}nlieb.
\newblock Simultaneous semantic and instance segmentation for colon nuclei
  identification and counting.
\newblock {\em arXiv preprint arXiv:2203.00157}, 2022.

\bibitem{liu2020psi}
Lihao Liu, Xiaowei Hu, Lei Zhu, Chi-Wing Fu, Jing Qin, and Pheng-Ann Heng.
\newblock $\psi$-net: Stacking densely convolutional lstms for sub-cortical
  brain structure segmentation.
\newblock {\em IEEE transactions on medical imaging}, 39(9):2806--2817, 2020.

\bibitem{liu2022pc}
Lihao Liu, Zhening Huang, Pietro Li{\`o}, Carola-Bibiane Sch{\"o}nlieb, and
  Angelica~I Aviles-Rivero.
\newblock Pc-swinmorph: Patch representation for unsupervised medical image
  registration and segmentation.
\newblock {\em arXiv preprint arXiv:2203.05684}, 2022.

\bibitem{loshchilov2017decoupled}
Ilya Loshchilov and Frank Hutter.
\newblock Decoupled weight decay regularization.
\newblock {\em arXiv preprint arXiv:1711.05101}, 2017.

\bibitem{lu2022video}
Xiao Lu, Yihong Cao, Sheng Liu, Chengjiang Long, Zipei Chen, Xuanyu Zhou, Yimin
  Yang, and Chunxia Xiao.
\newblock Video shadow detection via spatio-temporal interpolation consistency
  training.
\newblock In {\em Proceedings of the IEEE/CVF Conference on Computer Vision and
  Pattern Recognition}, pages 3116--3125, 2022.

\bibitem{lu2019see}
Xiankai Lu, Wenguan Wang, Chao Ma, Jianbing Shen, Ling Shao, and Fatih Porikli.
\newblock See more, know more: Unsupervised video object segmentation with
  co-attention siamese networks.
\newblock In {\em Proceedings of the IEEE/CVF conference on computer vision and
  pattern recognition}, pages 3623--3632, 2019.

\bibitem{oh2019video}
Seoung~Wug Oh, Joon-Young Lee, Ning Xu, and Seon~Joo Kim.
\newblock Video object segmentation using space-time memory networks.
\newblock In {\em Proceedings of the IEEE/CVF International Conference on
  Computer Vision}, pages 9226--9235, 2019.

\bibitem{oord2018representation}
Aaron van~den Oord, Yazhe Li, and Oriol Vinyals.
\newblock Representation learning with contrastive predictive coding.
\newblock {\em arXiv preprint arXiv:1807.03748}, 2018.

\bibitem{patrick2021keeping}
Mandela Patrick, Dylan Campbell, Yuki Asano, Ishan Misra, Florian Metze,
  Christoph Feichtenhofer, Andrea Vedaldi, and Jo{\~a}o~F Henriques.
\newblock Keeping your eye on the ball: Trajectory attention in video
  transformers.
\newblock {\em Advances in neural information processing systems},
  34:12493--12506, 2021.

\bibitem{song2018pyramid}
Hongmei Song, Wenguan Wang, Sanyuan Zhao, Jianbing Shen, and Kin-Man Lam.
\newblock Pyramid dilated deeper convlstm for video salient object detection.
\newblock In {\em Proceedings of the European conference on computer vision
  (ECCV)}, pages 715--731, 2018.

\bibitem{tran2015learning}
Du Tran, Lubomir Bourdev, Rob Fergus, Lorenzo Torresani, and Manohar Paluri.
\newblock Learning spatiotemporal features with 3d convolutional networks.
\newblock In {\em Proceedings of the IEEE international conference on computer
  vision}, pages 4489--4497, 2015.

\bibitem{vaswani2017attention}
Ashish Vaswani, Noam Shazeer, Niki Parmar, Jakob Uszkoreit, Llion Jones,
  Aidan~N Gomez, {\L}ukasz Kaiser, and Illia Polosukhin.
\newblock Attention is all you need.
\newblock {\em Advances in neural information processing systems}, 30, 2017.

\bibitem{verma2022interpolation}
Vikas Verma, Kenji Kawaguchi, Alex Lamb, Juho Kannala, Arno Solin, Yoshua
  Bengio, and David Lopez-Paz.
\newblock Interpolation consistency training for semi-supervised learning.
\newblock {\em Neural Networks}, 145:90--106, 2022.

\bibitem{voigtlaender2019feelvos}
Paul Voigtlaender, Yuning Chai, Florian Schroff, Hartwig Adam, Bastian Leibe,
  and Liang-Chieh Chen.
\newblock Feelvos: Fast end-to-end embedding learning for video object
  segmentation.
\newblock In {\em Proceedings of the IEEE/CVF Conference on Computer Vision and
  Pattern Recognition}, pages 9481--9490, 2019.

\bibitem{wang2021single}
Tianyu Wang, Xiaowei Hu, Chi-Wing Fu, and Pheng-Ann Heng.
\newblock Single-stage instance shadow detection with bidirectional relation
  learning.
\newblock In {\em Proceedings of the IEEE/CVF Conference on Computer Vision and
  Pattern Recognition}, pages 1--11, 2021.

\bibitem{wang2020instance}
Tianyu Wang, Xiaowei Hu, Qiong Wang, Pheng-Ann Heng, and Chi-Wing Fu.
\newblock Instance shadow detection.
\newblock In {\em Proceedings of the IEEE/CVF Conference on Computer Vision and
  Pattern Recognition}, pages 1880--1889, 2020.

\bibitem{wang2018non}
Xiaolong Wang, Ross Girshick, Abhinav Gupta, and Kaiming He.
\newblock Non-local neural networks.
\newblock In {\em Proceedings of the IEEE conference on computer vision and
  pattern recognition}, pages 7794--7803, 2018.

\bibitem{wolf2019huggingface}
Thomas Wolf, Lysandre Debut, Victor Sanh, Julien Chaumond, Clement Delangue,
  Anthony Moi, Pierric Cistac, Tim Rault, R{\'e}mi Louf, Morgan Funtowicz,
  et~al.
\newblock Huggingface's transformers: State-of-the-art natural language
  processing.
\newblock {\em arXiv preprint arXiv:1910.03771}, 2019.

\bibitem{xie2021segformer}
Enze Xie, Wenhai Wang, Zhiding Yu, Anima Anandkumar, Jose~M Alvarez, and Ping
  Luo.
\newblock Segformer: Simple and efficient design for semantic segmentation with
  transformers.
\newblock {\em Advances in Neural Information Processing Systems},
  34:12077--12090, 2021.

\bibitem{yan2022multiview}
Shen Yan, Xuehan Xiong, Anurag Arnab, Zhichao Lu, Mi Zhang, Chen Sun, and
  Cordelia Schmid.
\newblock Multiview transformers for video recognition.
\newblock In {\em Proceedings of the IEEE/CVF Conference on Computer Vision and
  Pattern Recognition}, pages 3333--3343, 2022.

\bibitem{zhang2021vidtr}
Yanyi Zhang, Xinyu Li, Chunhui Liu, Bing Shuai, Yi Zhu, Biagio Brattoli, Hao
  Chen, Ivan Marsic, and Joseph Tighe.
\newblock Vidtr: Video transformer without convolutions.
\newblock In {\em Proceedings of the IEEE/CVF International Conference on
  Computer Vision}, pages 13577--13587, 2021.

\bibitem{zhao2017pyramid}
Hengshuang Zhao, Jianping Shi, Xiaojuan Qi, Xiaogang Wang, and Jiaya Jia.
\newblock Pyramid scene parsing network.
\newblock In {\em Proceedings of the IEEE conference on computer vision and
  pattern recognition}, pages 2881--2890, 2017.

\bibitem{zheng2019distraction}
Quanlong Zheng, Xiaotian Qiao, Ying Cao, and Rynson~WH Lau.
\newblock Distraction-aware shadow detection.
\newblock In {\em Proceedings of the IEEE/CVF Conference on Computer Vision and
  Pattern Recognition}, pages 5167--5176, 2019.

\bibitem{zhou2017scene}
Bolei Zhou, Hang Zhao, Xavier Puig, Sanja Fidler, Adela Barriuso, and Antonio
  Torralba.
\newblock Scene parsing through ade20k dataset.
\newblock In {\em Proceedings of the IEEE Conference on Computer Vision and
  Pattern Recognition}, 2017.

\bibitem{zhou2019semantic}
Bolei Zhou, Hang Zhao, Xavier Puig, Tete Xiao, Sanja Fidler, Adela Barriuso,
  and Antonio Torralba.
\newblock Semantic understanding of scenes through the ade20k dataset.
\newblock {\em International Journal of Computer Vision}, 127(3):302--321,
  2019.

\bibitem{zhu2018bidirectional}
Lei Zhu, Zijun Deng, Xiaowei Hu, Chi-Wing Fu, Xuemiao Xu, Jing Qin, and
  Pheng-Ann Heng.
\newblock Bidirectional feature pyramid network with recurrent attention
  residual modules for shadow detection.
\newblock In {\em Proceedings of the European Conference on Computer Vision
  (ECCV)}, pages 121--136, 2018.

\end{thebibliography}
